\documentclass{llncs}

\usepackage{epsfig}

\begin{document}

\title{A philosophical essay on life and its connections
with genetic algorithms}

\author{Fernando G. Lobo}
\institute{
   ADEEC-FCT\\
   Universidade do Algarve\\
   Campus de Gambelas\\
   8000-062 Faro, Portugal\\
   flobo@ualg.pt
}

\maketitle

\begin{abstract}
This paper makes a number of connections between life
and various facets of genetic and evolutionary algorithms 
research. Specifically, it addresses the topics of adaptation,
multiobjective optimization, decision making, deception, and
search operators, among others.
It argues that human life, from birth to death, is an adaptive or 
dynamic optimization problem where people are continuously
searching for happiness.
More important, the paper speculates that genetic algorithms 
can be used as a source of inspiration for 
helping people make decisions in their everyday life.

\end{abstract}

\section{Introduction}
\label{sec:introduction}
Genetic algorithms (GAs) have been mainly used as a tool for 
search and optimization in a variety of problem domains. 
When they were invented however the original goal was not
to solve specific problems, but rather to understand adaptation
principles that occur in nature, and to find ways of incorporating
those principles in computer algorithms \cite{Holland:75a}.

At present time, search and optimization still receives the majority
of the attention, but the spectrum of GA applicability have gone well
beyond that.  GAs have been used for automatic 
programming \cite{Koza:99}, for understanding
biological systems \cite{CollinsRJ:92},
for modeling communities in ecologies \cite{Holland:95},
for modeling social organizations \cite{Kosorukoff2:2002:gecco}, 
and also as computational models of innovation and creativity 
\cite{Goldberg:2002}.

This paper discusses yet another facet; GAs as a computational
model for decision-making in life. It speculates that a person's
life follows a path that has striking similarities with the way 
genetic algorithms work. Moreover, it argues that many problems 
encountered in day-to-day life have a direct counterpart in some 
aspect of GA research.
At first sight, such provocative statement may seem exaggerated, 
but I hope to convince the reader that the argument is valid and 
is interesting enough to be known and shared with the evolutionary 
computation community. 

Genetic algorithms have been defined as search procedures based
on the mechanics of natural selection and genetics \cite{Goldberg:89d}.
Typical GAs consist of a generational
loop, where artificial individuals are born, survive, mate, and 
eventually die. The creation of individuals does not happen randomly.
Individuals inherit characteristics from their ancestors, and the 
fit ones propagate their traits to future generations. 
In the face of it, GAs can be thought of as a computer simulation of 
an evolutionary system, and in particular, as a simulation of life.

In this paper, however, life is not addressed in the context of an 
evolutionary process, generation after generation. Instead, it is 
addressed in the common sense of the word. It argues that a person's life, 
from birth to death, can be seen as an adaptation and/or optimization 
process, and is faced with many of the challenges that have been
investigated by GA researchers.

\section{Understanding life}
The best way that I found to set the stage for the rest of the paper,
it to start by quoting two Buddhist philosophers.

\begin{quote}
I believe that the very purpose of our life is to seek happiness. That is
clear. Whether one believes in religion or not, whether in this religion
or that religion, we are all seeking something better in life. So, I think,
the very motion of our life is towards happiness. 
(Dalai Lama, 1998)
\cite{DalaiLama:1998}
\end{quote}

\begin{quote}
Life has no real objective. The best objective is happiness. However,
happiness can be misunderstood easily. One best way to understand life
is to talk about it. Then we may be able to get rid of our ignorance 
and become wise about life. This is very important because life is so short.
(Hongsapan, 1990)
\cite{Hongsapan:1990}
\end{quote}

Similar paragraphs may be found in other sources, but the purpose
of this paper is not to talk about Buddhist philosophy
or religion. Such topics are out of the scope of a genetic 
and evolutionary computation conference.
The quotes however contain a number of elements that have
triggered my thinking into evolutionary computation.
In particular, the following three elements are worth emphasizing:

\begin{itemize}
\item life
\item objective
\item time
\end{itemize}

First and foremost, it is important not to forget
that one of the purposes of the research on evolutionary computation
is not only to solve artificial problems but also to get insights
about real living systems. Therefore, as GA researchers, it is 
important that we seek to understand life and it is important that
we talk (and write) about it. 
%That is the main reason why I am writing
%this paper (not to mention that writing this paper is fun and that makes
%me happy).

For several years I have had in my mind the idea of the existence
of a connection between optimization and life. To be honest, that
was one of the main reasons why I got attracted to genetic algorithms
in the first place. Independently of religious preferences, I believe
it is sensible to assume that most people would agree that the objective
of life is happiness. Apparently it is a hard problem because people never
seem to have enough of it.

Another important element of the quotation is the {\em time} issue.
Life is short and time is precious. Therefore, humans should not waste 
time in things that make them unhappy. When applying genetic algorithms
for optimization problems or for modeling adaptation, time is also a 
very important factor. In optimization, we are interested in
achieving the best possible solution as quick as possible. In adaptation
problems, we are interested in continuously adjust to a changing
environment, but time is important nonetheless because we want our 
system to be able to react quickly.

A number of other aspects stand out, each of which is better explored
in the remaining of the paper.
During the next sections, we will be discussing various aspects that
have been studied in the evolutionary computation field, making 
analogies of those aspects to things that happen in real life. 
Specifically, the paper addresses the topics of multi-criteria 
decision making, non-stationary environments, deception, operators,
and search strategies.

\section{The search space of life}
Human beings make decisions and take actions continuously, even if most
of them are done in an unconscious way.
These decisions or actions can be looked at from either a macro 
or  microscopic time scale. From a macroscopic point of view,
when we analyze our lives, we usually restrict ourselves to
the most important decisions that we have taken in the past or 
that we expect to do in the future. Examples include the decision 
to embark on doctoral studies, to work and live in a 
foreign country, to get married, to have children, and so on.
On a microscopic time scale, decisions are also made constantly. For example,
at this moment I am writing a paper about life and its connections with 
genetic algorithms, but I could have decided to go to a movie, practice 
some Yoga, read a book, chat with a friend, or do something else instead. 
Likewise, you as reader are reading this paper, but there are many other 
things that you could be doing instead.

The accumulation of all the decisions that we make through time
reflect in some way the type of person that we are. There are of course 
external factors that 
may influence what we are and what we will be, but the decisions and actions 
that we take during our life determine to some extent what we are today and
will determine what we will be in the future.

For the purpose of this paper, the important thing that I would like to
stress is that there are many different
ways in which we can live our lives. The number is so large that one
could say that the search space of life is unlimited.
When we are born we are faced with a whole world to explore, just
like an artificial individual at generation zero of a randomly initialized 
population.

\section{Life's objective is hard to quantify}
We have seen that life has a large search space and have 
started to see its connections with a typical run of a genetic algorithm.

Hongsapan's quote early in this paper started with a kind of contradiction.
On the one hand, he argues that life has no real objective. On the other
hand, he says that life's best objective is happiness. The contradiction
is interesting, and in my modest opinion, reveals the fact that happiness
is a non-measurable concept, something that is very hard to quantify.

Having a non-measurable objective is not a reason for crossing out
the analogies between life and genetic algorithms. As opposed to other
optimization methods, genetic and other methods of evolutionary computation
are able to handle objectives that are not measurable by a pure 
mathematical form. 
There are  examples of successful GA applications where the fitness 
of solutions are not obtained through a mathematical 
objective function, but rather through a subjective 
process. One such example is an application for tracking criminal 
suspects \cite{Caldwell:1991}. 
There the value of a solution is obtained by having 
a human assign a subjective score to it.
Likewise, in real life people can choose among different alternatives 
by subjectively choosing the one that leads to higher levels of happiness.

It is worth spending additional
time talking about happiness. Human beings live in society, and so, 
happiness is a concept that cannot be seen in isolation. In some ways
it is a kind of a recursive concept. Most human beings are happy when 
they can contribute to the happiness of others. Everybody seeks happiness, 
even those who are altruistic and apparently ``sacrifice'' their lives 
to a certain cause. Moreover, the decisions and actions that we take
in our life may have an effect that goes beyond our own lifetime.
Consider for example the work of a scientist, or an artist, or a 
politician. It is perfectly normal for a person to
sacrifice his/her life for the benefit of a cause, for the benefit 
of future generations, or for the benefit of humanity in general.
By doing so, that person feels happy.

It is interesting to observe that the problem of quantifying happiness 
appears to have similar characteristics to the problem of assessing the
value of an environmental resource. For example, what is the value of 
the Amazon forest? Researchers in the field of 
{\em Environmental Economics} have been studying this type of question
\cite{Daly:1997}, and its similarity with the 
problem of quantifying happiness is quite strong. 
Both are hard problems because their impact
spans over our own lifetime and prolongates to future generations.

\section{Happiness involves multiple objectives}
There is another reason why happiness is hard to quantify;
happiness means different things to different people. In some sense 
it is a concept that involves multiple objectives. 
People have different objectives or goals that
wish to be fulfilled in life. Examples may include getting rich, 
getting famous, having a successful career, living in a location 
with nice weather conditions, having a nice house, eating good food, 
spending time with the family, traveling around the world,
and so on.

These objectives may be conflicting sometimes and the difficult
thing is obtaining a balance between all the things 
that we would like to do. In the face of it, many of us end up 
making compromises because it is impossible to optimize each 
objective on its own with affecting the other ones. Using the parlance 
of multiobjective optimization, people move constantly along a trade-off 
solution surface of non-dominated solutions \cite{Deb:2001}, 
and then make choices based on higher level information (most of the time
just an intuition that one thinks will lead to higher levels of 
happiness).

In summary, life is full of decisions involving conflicting 
objectives. The same thing holds for many real world optimizations 
problems, and genetic algorithms have shown to be useful for that
purpose \cite{Deb:2001}. 

\section{Life occurs in a non-stationary environment}
For most optimization problems, the objective function is static. That is,
it does not change through time. Many problems however have a dynamic
objective function, a function that changes during the optimization
process itself. An example is the inclusion of new orders in a 
vehicle routing problem. There, it does not make sense to talk about 
a single best solution. Instead, the goal is to be able to continuously 
adapt and follow the best solution as closely as possible at any one time.

GAs that are used for static optimization problems have trouble with 
non-stationary functions. What happens is that after a period 
of time the population converges (or almost converges) to a single
solution. Once that happens, it is very hard to explore new solutions,
and thus, the algorithm fails to adapt to the changing environment. 
Fortunately, variations of standard GAs and other 
evolutionary algorithms have been investigated and engineered to be able
to handle this class of problems. The critical issue is to able to quickly
react to changes on the environment, something that can be done is several
ways such as using diversity preservation techniques, or using 
auxiliary memory \cite{Branke:2001}.
It is no coincidence that many researchers interested in dynamic 
optimization problems have approached evolutionary methods as 
a source of inspiration. After all, natural evolution is a
process of continuous adaptation.

Going back to our philosophical discussion about life, there 
should be no doubt that we are also dealing with a similar kind
of problem. Humans try to maximize happiness, but what makes us
happy today might not make us happy tomorrow. Our goal is to 
continuously seek happiness, day after day, year after year.
In order to do that, global convergence should be avoided 
by all means. Life has no optimal solution. The important thing
is to seek happiness and be prepared to react to changes in a timely
manner. Once again, research in genetic algorithms is teaching us
an important lesson. Humans should not be narrow-minded. We need to 
be open to different views, and never stop exploring.

\section{Life can have traps}
\begin{figure}
\centering
\epsfig{figure=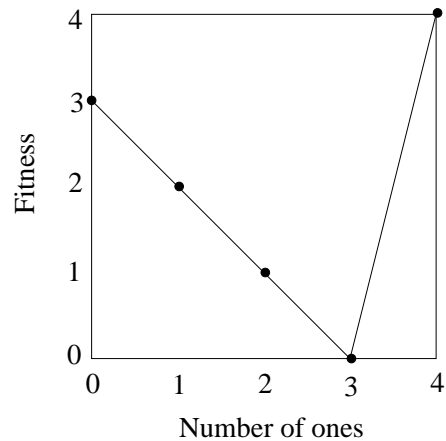,width=0.5\textwidth}
\caption{Example of a 4-bit trap function.}
\label{fig:trap}
\end{figure}

Deception has been introduced by \cite{Goldberg:87d} to test
the limitations of GAs. An example of a deceptive function
is the so-called trap function as shown in Figure~\ref{fig:trap}.
A trap function is a function of the number of ones in a bitstring. 
It is a problem that is difficult for GAs (and for any other method)
because it misleads the search to a solution that is far away from
the right answer. 
In the example shown in Figure~\ref{fig:trap} there is the tendency
for a solution to go to 0000 (with fitness 3), which is maximally
away from the best solution, 1111 (with fitness 4). The problem is
misleading because small incremental changes on a solution are rewarded,
leading the search to a deceptive attractor. Once there, it is always
possible to jump to the correct answer by continuing doing small
incremental changes \cite{Muhlenbein:92c*}.
Unfortunately, doing so is likely to take a very long time, 
and time is very important because life is short. Another quotation
is appropriate.

\begin{quote}
Time is funny. You might fully realize it when you had been too old
to help yourself. Life had passed so fast so you lied to yourself that
you had a long life. \cite{Hongsapan:1990}
\end{quote}

For many years, Goldberg and his students have tried to design GAs 
that could solve problems of bounded deception in a timely fashion. 
By bounded deception it is assumed 
a problem that may contain combinations of some bits that together
give rise to misleading solutions. 
In recent years, a number of algorithms have been developed and
proved to be competent
in solving such problems \cite{Goldberg:2002}.

Once more, the lessons learned for overcoming traps when solving
artificial problems, may be transferred to real life in order to
help people recognize the traps of life that one may encounter here
and there. Some may disagree, but I firmly believe that deception 
does occur in life. It typically happens in those situations when one 
is not capable of looking far ahead enough, and simply follows the 
easy path. Herein I argue that the important thing is that one 
is able to recognize a trap, and
once that happens, avoid by all means continuing doing small 
incremental changes. Instead, use GAs as a source of inspiration,
and try escaping traps in more efficient ways. For example, by 
enlarging the population of our past decisions and finding the 
necessary linkages to allow escaping the trap in a quick way.

Deception in real life is not confined to a person's life. I have
observed deception in organizations as well. Moreover, I have observed
that organizations fail many times to recognize traps, and 
insist on doing small incremental changes hoping that things will
improve. Examples include important political reforms, or changes 
in national educational systems.
It is my personal belief that genetic algorithms can be used as a
source of inspiration for making organizations respond
more effectively. Indeed,
research in that direction has already begun \cite{Kosorukoff2:2002:gecco}.

\section{GA operators in real life}
We have been discussing several topics that have been investigated
in GA research but so far have not said much about GA operators. 
Herein, we briefly discuss selection, mutation, and 
crossover, and its correspondence in real life.

\subsection{Selection}
Selection is the operator that distinguishes good from bad solutions.
It is something that humans do continuously in life. Again,
a quotation from the Dalai Lama seems appropriate.

\begin{quote}
Generally speaking, one begin by identifying those factors which lead
to happiness and those which lead to suffering. Having done this, one
then sets about gradually eliminating those factors which lead to suffering
and cultivating those which lead to happiness. \cite{DalaiLama:1998}
\end{quote}

The similarities with the selection operator of a GA are strong.
Humans seem to have a way of naturally eliminating things
that lead to suffering and keeping things that lead to happiness.
Next, we try to understand how humans explore new things.

\subsection{Variation: Mutation and Crossover}
If one observes the way most people live, one has to
conclude that, with some exceptions once in a while, what people
usually do in life is mostly small variations of what they have done 
in the past. In many ways, what we do today is not much different 
from what we did yesterday, and that sounds a lot like a mutation 
kind of exploration way of life.

Likewise, one could speculate that crossover corresponds to the big
decisions than one makes in life. Tracing back my own life, I can 
clearly recognize some landmarks that were very important for my 
development. In many ways, those were decisions that required a bit 
more thought, were more unusual, more challenging, more risky,
but were also the ones that yielded the highest payoff
on the long run.

One thing is definitely sure, people do not make decisions
randomly; life is not a random search.

\subsection{Early versus late generations}
It is interesting to observe what happens in a typical GA run and compare
it with real life. During the first generations the population contains 
a lot of diversity. Then as the search progresses, the individuals 
start following a kind of path and the number of exploration possibilities
have a tendency to diminish.

One could argue that the same thing also happens in real life. As children,
we are faced with a vast array of possibilities. We can grow up to be a 
researcher or grow up to be a policeman.
Then, as we grow up, we start making choices and the spectrum of things 
that we can continue to explore seem to get more confined and follow
a certain kind of path. I argue that since life is an adaptive process,
it is wise to be open-minded and not follow a particular path blindly.
Instead, it is important to continue exploring because otherwise there
is a risk of converging prematurely. If that happens, we may have a hard
time seeking happiness.

\section{Summary and conclusions}
\label{sec:conclusions}

When someone works for some time on a particular topic, there is
a tendency for that person to get deeply involved with it. 
Having worked with genetic algorithms for some years, I got to the point 
where I start seeing them everywhere, making analogies of its mechanisms
with things that in the first place seem unrelated to GAs.
I hope this is not an indication of insanity, and I have made my
best effort to articulate those connections and convince the reader that 
several aspects that have been studied in the field of genetic 
algorithms have a direct counterpart in real life.

This is an unusual kind of paper, in the sense that it is neither a 
theoretical study, nor an improvement or modification of an
existing algorithm, nor an application to benchmarking problems.
Instead, it is a speculative and more philosophical kind of paper.

The work presented herein may be extended to further developments which
may lead to a better understanding of life. I have talked about some of the
analogies that exist between genetic algorithms and life, but I am sure
that additional analogies can be made. 
An interesting thing to explore is to understand how the work on 
anticipatory systems \cite{Rosen:1985} \cite{Butz:2002} fits in here. 
These systems are based on psychological theories that say that
animal and human behavior is influenced by anticipations.
In other words, when someone faces a number of alternatives and
has to choose one, the decision is largely made by anticipating
(or predicting) the future under each alternative
and then choosing the one that appears to give a better reward.
This is in some ways the opposite of the way genetic algorithms work.
In GAs, new solutions are explored based on the past experiences
of the algorithm, not through anticipation. However, one could also
argue that anticipatory systems need to have knowledge about the
past in order to make good anticipations. We can only anticipate the
future due to our knowledge of past experiences. 

As a final remark, it is interesting to observe that the pioneers of the field 
of genetic and evolutionary algorithms have used nature and life as an
inspiration for designing computer algorithms. This paper
speculates that the reverse is also true; that genetic 
algorithms can be used as a source of inspiration for life itself.

\section*{Acknowledgements} 
I thank my brother, Gon\c{c}alo Lobo, for his comments on some of the issues
raised in this paper.

This work was sponsored in part by the Portuguese Foundation for Science
and Technology (FCT/MCT), under grant
POSI/SRI/42065/2001 and grant POCTI/MGS/37970/2001.

\bibliographystyle{splncs}
\bibliography{references}

\end{document}